\documentclass{article}
\usepackage{spconf,amsmath,graphicx}
\usepackage{comment}
\usepackage{amsfonts}
\usepackage{graphicx}
\usepackage{multirow}
\usepackage{booktabs}
\usepackage{cleveref}
\usepackage{graphicx}
\usepackage{subcaption}

\title{Low-Rank Embedding of Kernels in Convolutional Neural Networks\\ under Random Shuffling}
\name{$^1$Chao Li, $^1$Zhun Sun, $^{1,2}$Jinshi Yu, $^1$Ming Hou and $^1$Qibin Zhao\thanks{Corresponding Email: \{chao.li, qibin.zhao\}@riken.jp}}
\address{$^1$RIKEN Center for Advanced Intelligence Project (AIP), Tokyo 103-0027, Japan\\
        $^2$School of Automation, Guangdong University of Technology, Guangzhou 510006, China}
\begin{document}
%
\maketitle
\begin{abstract}
Although the convolutional neural networks (CNNs) have become popular for various image processing and computer vision tasks recently, it remains a challenging problem to reduce the storage cost of the parameters for resource-limited platforms.
In the previous studies, tensor decomposition (TD) has achieved promising compression performance by embedding the kernel of a convolutional layer into a low-rank subspace. 
However the employment of TD is naively on the kernel or its specified variants.
Unlike the conventional approaches, this paper shows that the kernel can be embedded into more general or even random low-rank subspaces.
We demonstrate this by compressing the convolutional layers via \emph{randomly-shuffled} tensor decomposition (RsTD) for a standard classification task using CIFAR-10. In addition, we analyze how the spatial similarity of the training data influences the low-rank structure of the kernels.
The experimental results show that the CNN can be significantly compressed even if the kernels are randomly shuffled. Furthermore, the RsTD-based method yields more stable classification accuracy than the conventional TD-based methods in a large range of compression ratios.
\end{abstract}


\section{Introduction}
\label{sec:intro}
Deep convolutional neural networks (CNNs) have advanced to show the state-of-the-art performance in image processing and computer vision applications~\cite{ulyanov2018deep,ren2015faster,dong2014learning}.
However, the huge storage cost of the trainable parameters severely limits its deployment in practice, especially on resource-limited platforms such as smart-phones and wearable devices. For example, the AlexNet Caffemodel is over 200MB and the VGG-16 Caffemodel is over 500MB~\cite{han2015deep}.
Thence, how to efficiently compress the parameters of CNNs has become an urgent task and challenging problem.

To address this problem, many methods have been proposed, including encoding, quantization and pruning~\cite{han2015deep}. 
In recent studies, the deep neural networks are also compressed by the tensor decomposition (TD) model, that embeds a multi-way array into a lower dimensional space~\cite{cichocki2015tensor}. 
TD itself shows promising results with high compression ratio with less degraded performance~\cite{lebedev2014speeding,novikov2015tensorizing,stoudenmire2016supervised,kossaifi2017tensor,yang2016deep,kim2015compression}. Besides, a combination of TD and aforementioned methods, \textit{e.g.}, encoding and pruning, can further improve the compactness of the CNNs~\cite{anonymous2019exploiting}.
In early studies, TD was directly employed on the learned kernels. For example, Lebedev \textit{et al.} exploited CANDECOMP/PARAFAC decomposition to compress the convolutional layers in AlexNet~\cite{lebedev2014speeding}. Similarly, Kim \textit{et al.} used Tucker decomposition to speed up various CNNs~\cite{kim2015compression}. More recently, more sophisticated TD models such as tensor train and tensor ring were also employed on compression, in which the kernels are tensorized into a higher-order tensors for higher compression level~\cite{garipov2016ultimate,wang2018wide}.

It is worthwhile to mention that, these conventional TD-based compression methods depend on the occurrence of spatial or channel-wised linear dependence within the kernel, thus the kernel can be naturally embedded into a low-rank subspace with lower dimension. Meanwhile, such linear dependence is considered to be occasioned by the spatial similarity of the training data~\cite{denil2013predicting}.
Such insight give rise to the following questions:  Is the spatial similarity the essential cause of the low-rank structure of the kernels? In addition, how does the spatial similarity of the training data influence the low-rank structure the kernels?


\begin{figure}
    \centering
    \begin{subfigure}[b]{0.23\textwidth}
        \includegraphics[width=1\textwidth]{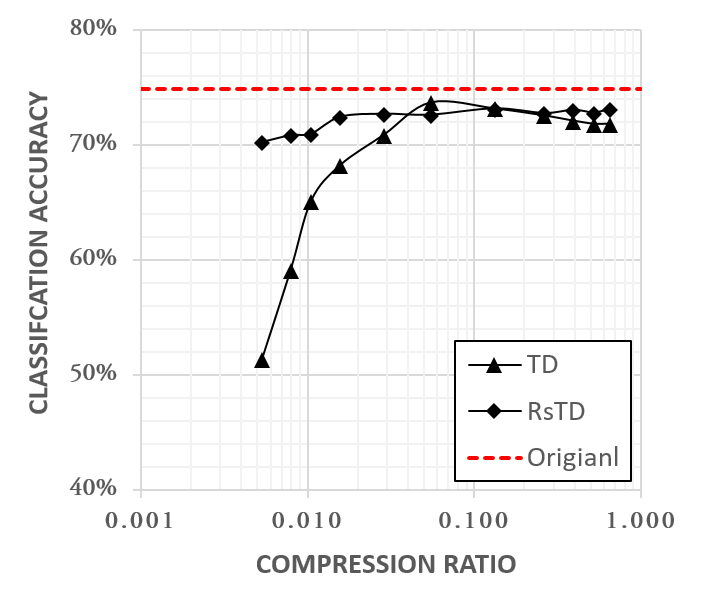}
        \caption{CIFAR-10}
        \label{fig:gull}
    \end{subfigure}
    \begin{subfigure}[b]{0.23\textwidth}
        \includegraphics[width=1\textwidth]{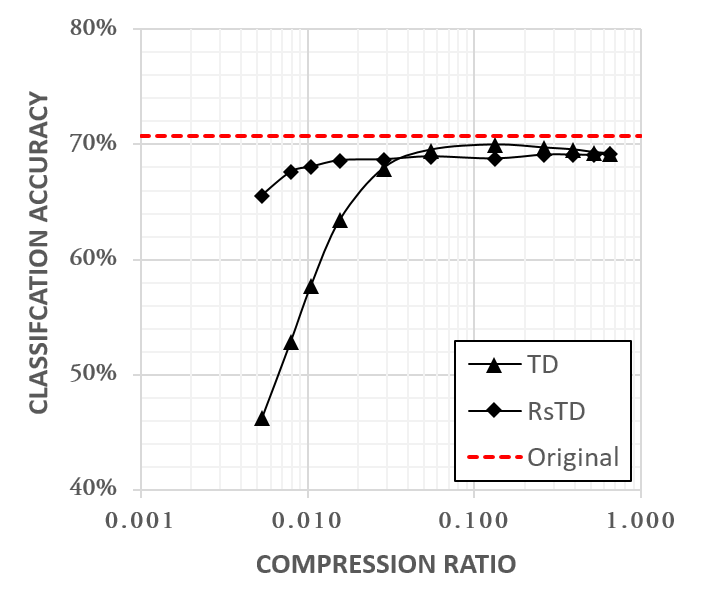}
        \caption{Noised CIFAR-10}
        \label{fig:tiger}
    \end{subfigure}
    \caption{Comparison of the classification accuracy of the CNNs in our experiments, where TD represents the conventional TD-based compression method (by tensor-train-matrix decomposition), RsTD denotes the proposed model in which the random shuffling operation is imposed on each kernel before TD, and the right line in the figure is the baseline by the uncompressed network.}\label{fig:example}
\end{figure}
To attempt to answer the questions above, we remove the influence of the spatial similarity by leveraging random shuffling the parameters in the kernels of convolutional layer before using TD (see \Cref{sec:Exp} for more details).
The experimental results reveal an interesting phenomenon that \textbf{the TD methods are able to compress CNNs effectively regardless of the random shuffling} (see \Cref{fig:example} for example), which implies the fact that the spatial similarity of the training data is not the key factor for embedding the kernels into a low-rank subspace.

Below, we first introduce an unified model for tensor decomposition by using the framework of tensor network. Using this model, we then propose the randomly-shuffled tensor decomposition (RsTD) based convolutional layer, which is used for CNN compression in Section \ref{sec:Rand}. After that, the experiments on two types of CNNs, namely TD-based and RsTD-based networks, by using CIFAR-10 dataset follow in \Cref{sec:Exp}. We conclude with the summary in Section \ref{sec:concl}.

\section{Formulation}
\label{sec:Rand}
\subsection{Unified model of tensor decomposition}
\begin{figure}
    \centering
    \includegraphics[width=0.33\textwidth]{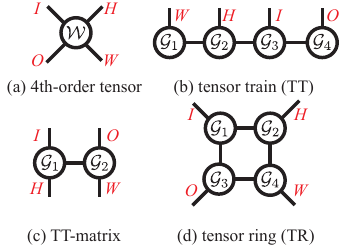}
    \caption{Graphical representation for decomposing a kernel ($4$th-order tensor) by using tensor train (TT), TT-matrix and tensor ring (TR) decomposition, respectively.}\label{fig:tensor_network}
    \vspace{-0.5cm}
\end{figure}

Tensors, also known as multi-way arrays, are generalization of the $2$nd-order matrices. 
Assuming the kernel of a convolutional layer as a $4$th-order tensor $\mathcal{W}\in\mathbb{R}^{I\times{}H\times{}W{}\times{}O}$, where $H,W$ denote the height and width of the filter and $I,O$ denote the number of input and output channels. Then a tensor decomposition model represents it as the multiplication of multiple latent core tensors. 
Under the framework of tensor network~\cite{cichocki2015tensor}, we can mathematically describe TD with a unified model, \textit{i.e.}
\begin{equation}
    \mathcal{W}=T_\mathbf{A}\left(\mathcal{G}_1,\mathcal{G}_2,\ldots,\mathcal{G}_N\right),
\end{equation}
where $\mathcal{G}_i,i\in\{1,2,\ldots,N\}$ denotes the latent core tensors and the operator $T_\mathbf{A}$ represents tensor multiplication with given adjacency matrix $\mathbf{A}$ that describe the graph structure of the TD model. For instance, the kernel $\mathcal{W}$ can be decomposed into four cores by tensor train (TT)~\cite{oseledets2011tensor} and the corresponding adjacency matrix is written as
\begin{equation}
    \mathbf{A}_{TT}=
    \left[
 \begin{matrix}
   0 & 1 & 0 & 0 \\
   1 & 0 & 1 & 0 \\
   0 & 1 & 0 & 1 \\
   0 & 0 & 1 & 0 
  \end{matrix}
  \right],\label{eq:A_TT}
\end{equation}
where rows and columns of \eqref{eq:A_TT} correspond different core tensors. We can also describe TD by using graphical representations, in which we consider the vertices as core tensors and the edges as the multiplication of two tensors. 
Fig. \ref{fig:tensor_network} shows three types of TD models including TT, TT-matrix and tensor ring  (TR)~\cite{zhao2016tensor}, all of which are used in the experiments in this paper. Due to the paper limit, more details about TD and its graphical representation can be found from~\cite{cichocki2016tensor} and the references therein.

\subsection{Randomly-shuffled TD (RsTD) Layer}
The conventional TD-based compression is a method to use the TD models to replace the original kernels in CNN.
In this paper, in order to embed the kernel into more diverse low-rank subspaces than the conventional method, we modify the conventional TD-based compression methods by imposing random shuffling operations.
Random shuffling operator is defined as a linear operator $R:\mathbb{R}^{W\times{}H\times{}I\times{}O}\rightarrow\mathbb{R}^{W\times{}H\times{}I\times{}O}$ which \emph{randomly} re-assigns a subscript index for each parameter of the kernel tensor $\mathcal{W}$, such that the parameters of the kernel will be relocated into another place.
To simulate the random characteristic by using algorithm, the mapping rule of $R$ can be determined by Fisher-Yates algorithm~\cite{batagelj2005efficient} or its modification~\cite{durstenfeld1964algorithm}, which gives equally likely permutation results.

By the random-shuffling operator $R$, the kernel is decomposed as 
\begin{equation}
    \Tilde{\mathcal{W}}=R\cdot{}T_{\mathbf{A}}\left(\mathcal{G}_1,\mathcal{G}_2,\ldots,\mathcal{G}_N\right).\label{eq:RsTD}
\end{equation}
In contrast to the conventional TD, the kernel generated by \eqref{eq:RsTD} will \emph{NOT} have the low-rank structure due to the random shuffling of the parameters. 
Even though the spatial similarity of the training data yields the low-rank kernels, the newly imposed random shuffling operator is able to invalidate this property.
%
The randomly-shuffled tensor decomposition (RsTD) based convolutional layer is constructed on the basis of \eqref{eq:RsTD} with the core tensors $\mathcal{G}_1,\mathcal{G}_2,\ldots,\mathcal{G}_N$ replacing the original kernel $\mathcal{W}$.
Mathematically, it can be formulated as
\begin{equation}
    \mathcal{Y}=f\left(\left[R\cdot{}T_{\mathbf{A}}\left(\mathcal{G}_1,\mathcal{G}_2,\ldots,\mathcal{G}_N\right)\right]\otimes{}\mathcal{X}+\mathcal{B}\right), \label{eq:layer}
\end{equation}
where the tensor $\mathcal{X}$ and $\mathcal{Y}$ denotes the input and output features of the layer, $f$ denotes the element-wise activation function, $\mathcal{B}$ denotes the bias and $\otimes{}$ represents the convolution operation\footnote{Here we ignore the stride and padding for brevity.}. 
 The randomness of R enables the kernels to be embedded into different low-rank subspaces with different training procedure. 
Furthermore, note that \Cref{eq:layer} can be degenerated as the conventional TD-based layer if $R$ equals a identical mapping. It implies that the RsTD-based layer is a more general model than the conventional TD-based layer.


\section{Experimental results and analysis}
\label{sec:Exp}
\subsection{Experiment setting}
\begin{table}[t]
  \centering
  \begin{tabular}{c}
    \toprule
    \textbf{Input} \\
    \midrule
    conv -- $3\times 3$ -- 256 -- stride 1\\
    conv -- $3\times 3$ -- 256 -- stride 1\\
    conv -- $3\times 3$ -- 256 -- stride 2\\
    \midrule
    conv -- $3\times 3$ -- 256 -- stride 1\\
    conv -- $3\times 3$ -- 256 -- stride 1\\
    conv -- $3\times 3$ -- 256 -- stride 2\\
    \midrule
    conv -- $3\times 3$ -- 256 -- stride 1\\
    global average pooling \\
    fully connected-10 \\
    \midrule
    soft-max classifier \\
    \bottomrule
  \end{tabular}
  \caption{CNN configurations. The convolution layer parameters are denoted by conv --\textless kernel size\textgreater --\textless number of output channels\textgreater--\textless stride option\textgreater. }
  \label{table:structure}
\end{table}
To evaluate the compression capacity of TD and RsTD-based layers, we construct the CNN by the two types of layers for a standard classification task on the CIFAR-10 dataset~\cite{krizhevsky2009learning}.
Specifically, we build a prototype CNN with 8 convolutional layers followed by batch normalization~\cite{ioffe2015batch} and the ReLU activation function. 
A fully-connected layer is attached at the top of network, whose outputs are fed into a soft-max classifier. 
The detailed structure is provided in \Cref{table:structure}. 
Instead of the conventional CNN, we compress the kernels from the 2nd to the final convolutional layer in the experiment by using TD and RsTD, respectively. Meanwhile, we choose TT-matrix, TT and TR as the decomposition model (see \Cref{fig:tensor_network}), and configure different ranks for each TD model to control the compression level. 

During training, we directly update the latent core tensors for each compressed layer using the stochastic gradient decent algorithm with a nesterov moment of 0.9. The initial learning rate is set to be 0.1, decayed by a factor of 10 at epoch 80 and 110, and the training stops at epoch 120. We repeat the training 5 times for each configuration and report the averaged classification accuracy.

Besides the original CIFAR-10 dataset, we further implement 2 variants of CIFAR-10 that are distorted by additional white Gaussian noise (AWGN) of standard deviation 0.4 and 0.8, respectively. Because imposing AWGN on the training data can decrease the spatial similarity for a image, we can leverage such the property to analyze how the spatial similarity of the training data influences the low-rank structure of the kernels. 
In this paper, we use the compression ratio to quantify the compression level for different models. Its formula is given by
\begin{equation}
    r_c=\frac{N_c}{N_u},
\end{equation}
where $N_c$ and $N_u$ denote the number of the parameters in the compressed and uncompressed networks, respectively.

\subsection{Results and analysis}
\begin{figure}[t]
    \centering
    \begin{subfigure}[b]{0.23\textwidth}
        \includegraphics[width=1\textwidth]{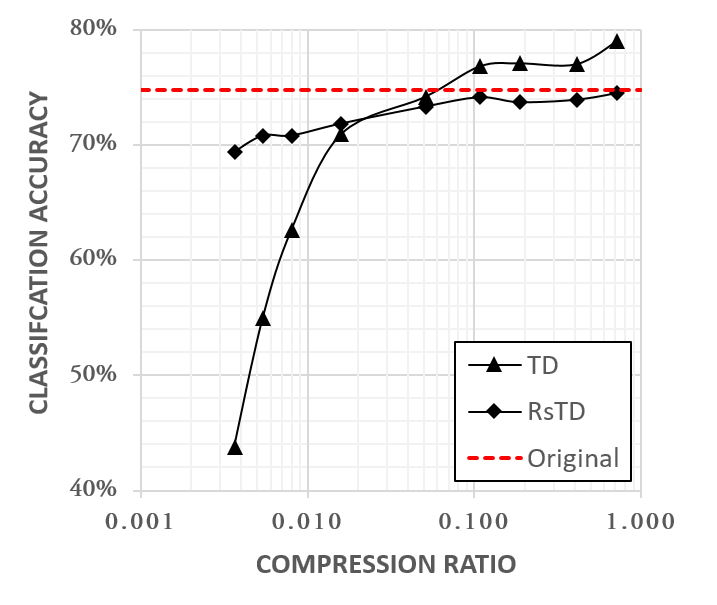}
        \caption{TT}
    \end{subfigure}
    ~ 
    \begin{subfigure}[b]{0.23\textwidth}
        \includegraphics[width=1\textwidth]{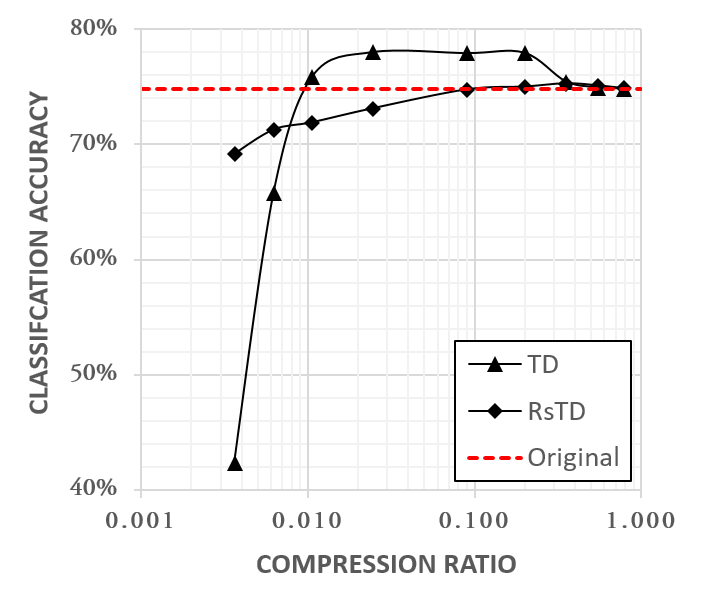}
        \caption{TR}
    \end{subfigure}
    \caption{Comparison of the test accuracy of CNNs with different compressed layers (TD and RsTD, respectively) by using the original CIFAR-10 dataset. In the figures, the red line denotes the uncompressed baseline, and different sub-figures represent different tensor decomposition models (TT and TR, respectively).}\label{fig:clean}
\end{figure}
\begin{figure}[t]
    \centering
    \begin{subfigure}[b]{0.23\textwidth}
        \includegraphics[width=1\textwidth]{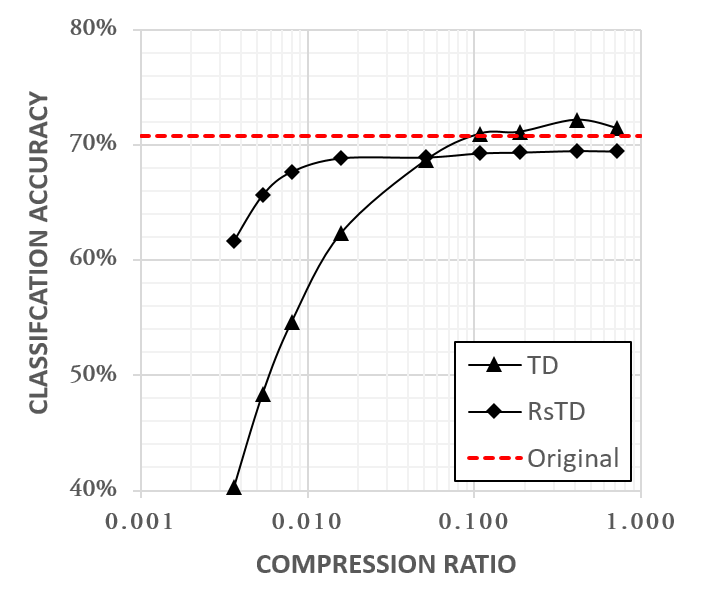}
        \caption{TT}
    \end{subfigure}
    ~ 
    \begin{subfigure}[b]{0.23\textwidth}
        \includegraphics[width=1\textwidth]{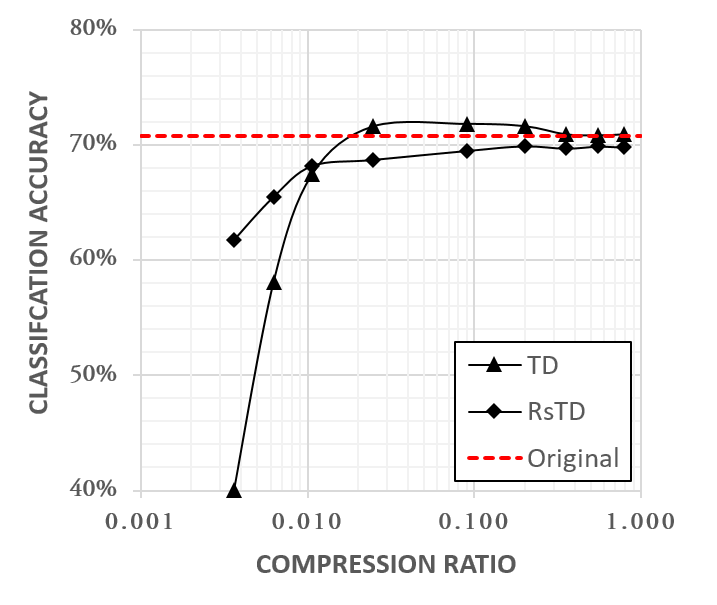}
        \caption{TR}
    \end{subfigure}
    \caption{Comparison of the test accuracy of CNNs with different compressed layers (TD and RsTD, respectively) by using the noised CIFAR-10 dataset (dev=0.4). In the figures, the red line denotes the uncompressed baseline, and different sub-figures represent different tensor decomposition models (TT and TR, respectively).}\label{fig:noise_0.4}
\end{figure}
\begin{figure*}[th]
    \centering
     \begin{subfigure}[b]{0.3\textwidth}
         \includegraphics[width=0.9\textwidth]{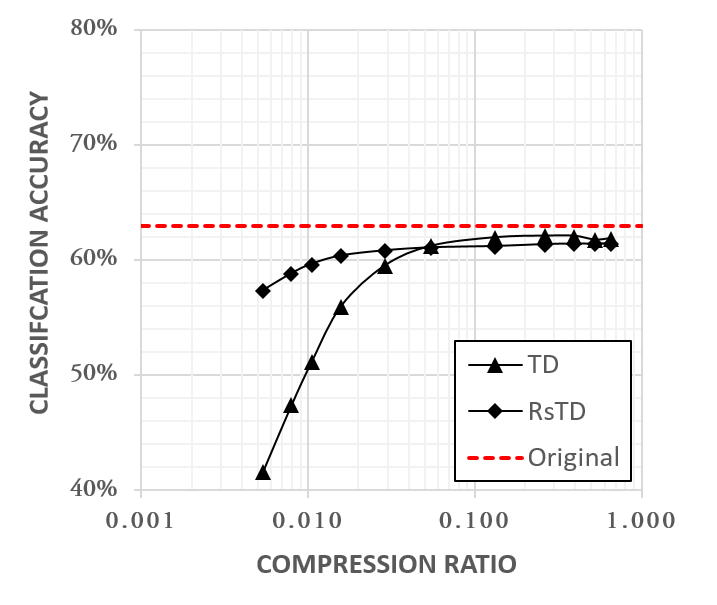}
         \caption{TT-matrix}
     \end{subfigure}
     ~ 
    \begin{subfigure}[b]{0.3\textwidth}
        \includegraphics[width=0.9\textwidth]{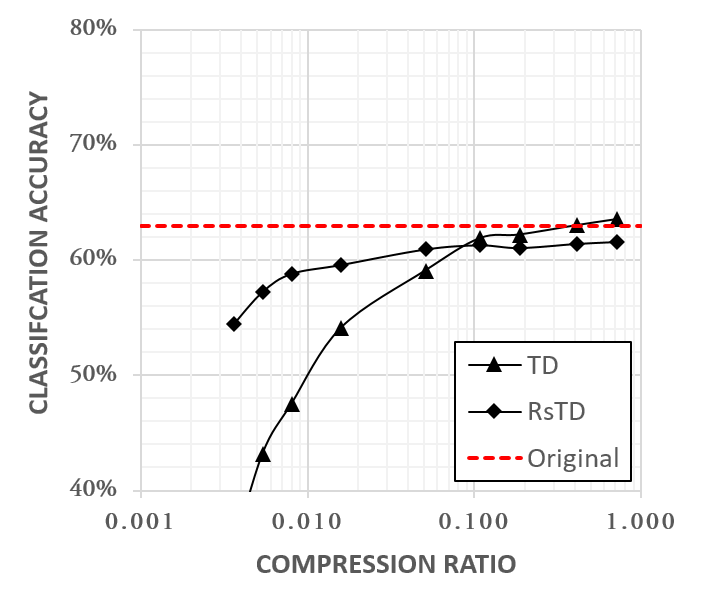}
        \caption{TT}
    \end{subfigure}
    ~ 
    \begin{subfigure}[b]{0.3\textwidth}
        \includegraphics[width=0.9\textwidth]{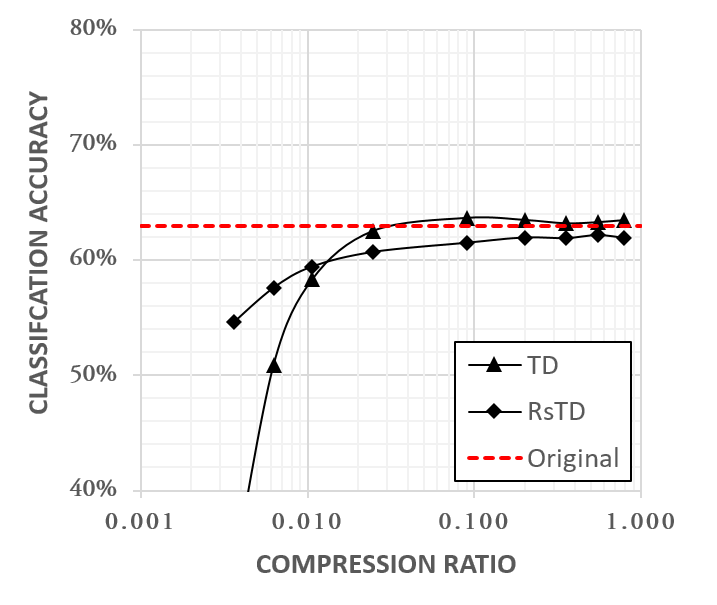}
        \caption{TR}
    \end{subfigure}
    \caption{Comparison of the test accuracy of CNNs with different compressed layers (TD and RsTD, respectively) by using the noised CIFAR-10 dataset (dev=0.8). In the figures, the red line denotes the uncompressed baseline, and different sub-figures represent different tensor decomposition models (TT and TR, respectively).}\label{fig:noise_0.8}
    \vspace{-0.3cm}
\end{figure*}
We first evaluate the classification performance of the compressed networks by using the original CIFAR-10.
\Cref{fig:example} (a) and \Cref{fig:clean} illustrate the classification accuracy of the trained CNNs on different tensor decomposition models with compression ratio ranging from 0.003 to 1.
As shown in the figures, both TD and RsTD-based layers obtain similar classification accuracy compared to the baseline (uncompressed networks, the red line in the figures) when the compression ratio is relatively high ($r_c>0.05{}$).
It implies that the storage overhead of the compressed CNN is 20 times smaller than the uncompressed counterpart by using both TD and RsTD-based layers without significant performance loss.
The accuracy of RsTD is competitive with that of baseline, which reflects the fact that the shuffled kernels can be still embedded into low-rank subspaces, even if the mapping rule $R$ is randomly chosen.

The performance curves in these figures render two interesting phenomena.
First, the accuracy of the TD-based layers go down dramatically as the compression ratio decreases, whereas the RsTD-based layers is able to maintain relatively high classification accuracy.  
The inferior performance of the TD-based layers is due to the under-fitting problem of the network. For example, assume that we decompose the kernel by using rank-1 TR decomposition ($r_c\approx0.004$), then Eq. \eqref{eq:layer} can be rewritten as
\begin{equation}
    \begin{split}
        &=f\left(\mathcal{G}_3\times_{1,3}\left(merge\left(\mathcal{G}_1,\mathcal{G}_2,\mathcal{G}_4\right)\otimes{}\mathcal{X}\right)+\mathcal{B}\right)
    \end{split},\label{eq:example}
\end{equation}
where the operators $\times_{1,3}$ and $merge(\cdot)$ denote the multiplication and merging operation of the core tensors~\cite{wang2018wide}, respectively. We can see from \eqref{eq:example} that the operations in the convolutional layer can be split into two steps. In the first step, the input feature $\mathcal{X}$ is convolved with the merged cores $\mathcal{G}_i,i=1,2,4$, which is equivalent to mapping $\mathcal{X}$ into a latent space. After that, the multiplication of the convolution output with $\mathcal{G}_3$, produces the features for each output channel by mixing the ``latent'' features. Hence, the rank of the TR model equals $1$ implies that the features for all output channels are identical to each other up to scale, and this fact naturally leads to under-fitting problem in most of CNN learning models. In contrast, the random shuffling operation of the RsTD-based layers will increase the rank of the kernel, which suggests the dimension of the ``latent'' features will also goes up. Such property can actually mitigate the under-fitting issue of the network.



The second phenomenon is that TT and TR-based layers outperform both the baseline and RsTD-based layers in the case of high compression ratio. This observation supports the conventional claim that the spatial similarity of the training data results in the linear dependence within the kernel. To find out the reason behind the performance gap between TD and RsTD-based layers, we validate our model on the noisy CIFAR-10 ($\mbox{dev}=0.4$ and 0.8, respectively) to train the networks. \Cref{fig:example} (b) and \Cref{fig:noise_0.4} give the classification accuracy when $\mbox{dev}=0.4$, and \Cref{fig:noise_0.8} illustrate the result when $\mbox{dev}=0.8$. As depicted in the figures, the performance gap between TD and RsTD becomes smaller as the noise strength increases. This is because imposing the noise will decrease the spatial similarity of the training data. In these cases, the RsTD-based layer shows more reliable performance with a larger range of compression ratios.

\section{Conclusion}
\label{sec:concl}
In this paper, we consider embedding the kernels in CNN into more general low-rank subspaces and analyze the impact of spatial similarity of the training data on the low-rank structure of the kernel. For this purpose, we impose the random shuffling operations before tensor decomposition. 
%
The experimental results demonstrate that RsTD can be exploited to compress the kernels in CNN without significant performance loss.
Also, CNNs equipped with RsTD-based layers overwhelmingly outperform those with un-shuffled kernels under significantly small compression ratio.
Furthermore, decreasing the spatial similarity of the training data diminishes the diversity of performance between TD and RsTD based models.
%
%
The conclusions can be made from the experimental results that the kernels in CNN have an inherent low-rank structure regardless of the structure of the training data.

\bibliographystyle{IEEEbib}
\bibliography{Template.bib}

\end{document}